\documentclass[letterpaper]{article} 
\usepackage{aaai2026}  
\usepackage{times}  
\usepackage{helvet}  
\usepackage{courier}  
\usepackage[hyphens]{url}  
\usepackage{graphicx} 
\usepackage{natbib}  
\usepackage{caption} 
\usepackage{algorithm}
\usepackage{algorithmic}

\usepackage{amsmath, amssymb, amsfonts}
\usepackage{adjustbox}  
\usepackage{array} 
\usepackage{bm}
\usepackage{booktabs}
\usepackage{makecell}
\usepackage{mathtools}
\usepackage{multirow}
\usepackage{tabularx}
\usepackage[table,xcdraw]{xcolor}
\usepackage{pifont}
\usepackage{bbold}
\usepackage{multicol}
\usepackage{microtype} 
\allowdisplaybreaks

\usepackage{newfloat}
\usepackage{listings}
\DeclareCaptionStyle{ruled}{labelfont=normalfont,labelsep=colon,strut=off} 
\floatstyle{ruled}
\newfloat{listing}{tb}{lst}{}
\floatname{listing}{Listing}
\title{Emotion-Coherent Reasoning for Multimodal LLMs\\ via Emotional Rationale Verifier}
\author{
    Hyeongseop Rha, 
    Jeong Hun Yeo, 
    Yeonju Kim, 
    Yong Man Ro\thanks{Corresponding author.}
}
\affiliations{
    Integrated Vision and Language Lab, KAIST, South Korea\\

    \{ryool\_1832, sedne246, yeonju7.kim, ymro\}@kaist.ac.kr
}

\begin{document}

\maketitle

\begin{abstract}
The recent advancement of Multimodal Large Language Models (MLLMs) is transforming human-computer interaction (HCI) from surface-level exchanges into more nuanced and emotionally intelligent communication. To realize this shift, emotion understanding becomes essential allowing systems to capture subtle cues underlying user intent. Furthermore, providing faithful explanations for predicted emotions is crucial to ensure interpretability and build user trust. However, current MLLM-based methods often generate emotion explanations that diverge from the target labels and sometimes even contradict their own predicted emotions. This inconsistency poses a critical risk for misunderstanding and erodes reliability in interactive settings. To address this, we propose a novel approach: the Emotional Rationale Verifier (ERV) and an Explanation Reward. Our method guides the model to produce reasoning that is explicitly consistent with the target emotion during multimodal emotion recognition without modifying the model architecture or requiring additional paired video–description annotations. Our method significantly improves faithful explanation–prediction consistency and explanation emotion accuracy on the MAFW and DFEW datasets. 
Through extensive experiments and human evaluations, we show that our approach not only enhances alignment between explanation and prediction but also empowers MLLMs to deliver emotionally coherent, trustworthy interactions, marking a key step toward truly human-like HCI systems.
\end{abstract}

\begin{links}
    \link{Code}{https://github.com/Rhatanii/ERV}
    \link{Extended version}{https://arxiv.org/abs/2510.23506}
\end{links}

\section{Introduction}
\begin{figure*}[t]
    \centering
    \includegraphics[width=1.0\textwidth]{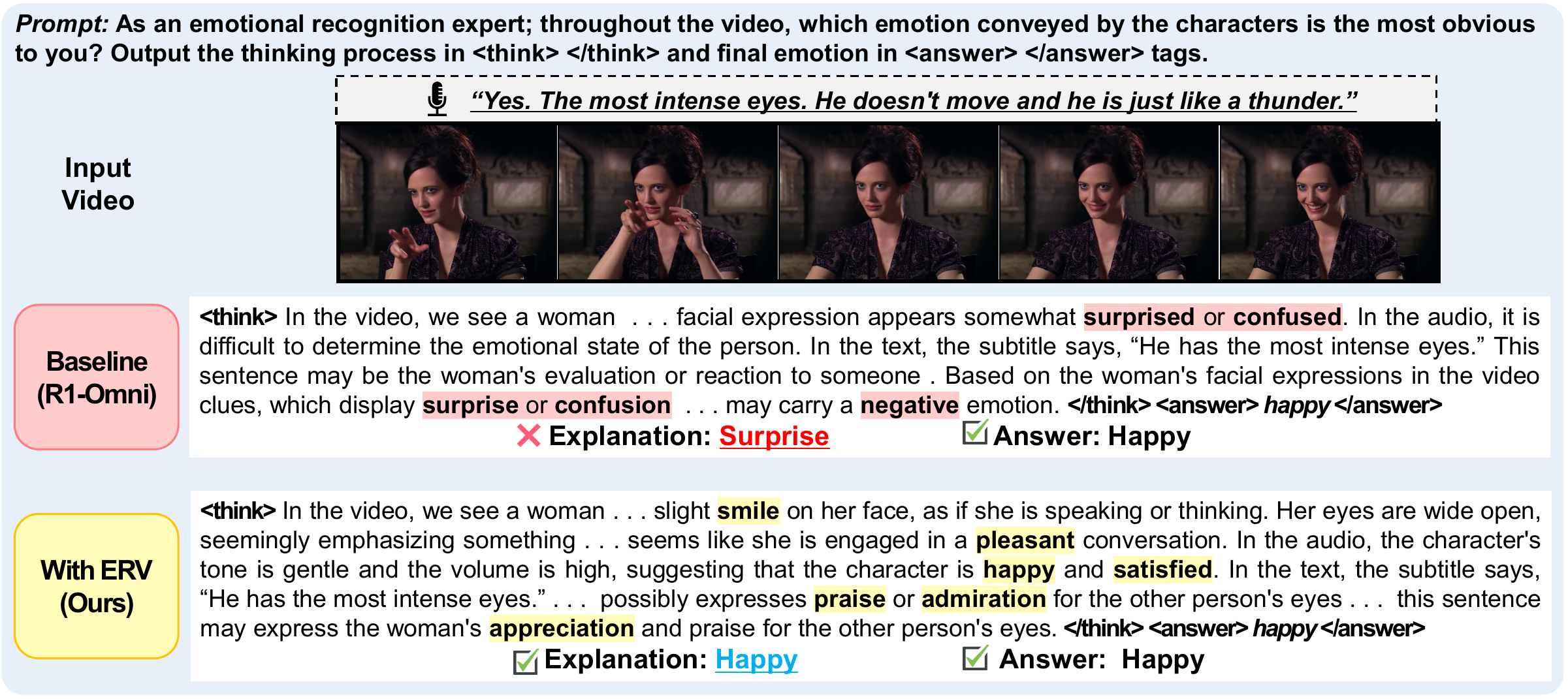}
    \caption{Illustrative comparison between the baseline and our model on explanation–emotion alignment (text truncated for brevity). While both correctly predict the final emotion, the baseline fails to generate an emotionally coherent explanation, highlighting its misalignment.}
    \label{fig:fig1}
\end{figure*}
As multimodal large language models (MLLMs)~\cite{chen2024internvl,li2024llava,wang2024qwen2vl, chu2024qwen2,cheng2024videollama} continue to evolve, they are reshaping human-computer interaction (HCI) by enabling systems to interpret cross-modal inputs (face video and speech audio), perform contextual reasoning, and generate human-like responses~\cite{fei2024empathyear,park2024let,kim2025speaking}. To support truly natural and emotionally aware communication, such systems must go beyond surface-level responses and capture the nuance, intent, and emotional context of user inputs. Among these capabilities, emotion understanding plays a central role in facilitating deeper, more meaningful interaction—particularly in applications such as psychological counseling~\cite{schmidgall2024agentclinic,lee2024cactus}, educational guidance~\cite{schutz2007emotion}, and empathetic dialogue systems~\cite{fei2024empathyear,zhang2025towards}.

While conventional multimodal emotion recognition (MER) approaches have treated emotion understanding as a classification task~\cite{zhang2024traditionalmer}, single-label predictions often fail to capture the complexity of human affect conveyed through face video and speech audio. To address this limitation, recent work has turned to \textit{descriptive emotion understanding}, leveraging MLLMs to generate natural language explanations that justify predicted emotions~\cite{lian2023EMER,cheng2024MERR,lianaffectgpt25, yang2025omni}. This shift is especially valuable in HCI scenarios, where users not only expect accurate emotional classification from visual and auditory cues but also seek coherent and human-like justifications to build trust and ensure transparency.

To enhance emotion reasoning, a recent study leverages Reinforcement Learning (RL) and Chain-of-Thought (CoT) prompting to train MLLMs~\cite{zhao2025r1}. Although this method improves classification accuracy, it fails to consistently produce emotionally coherent explanations. This limitation may be attributable to the datasets used in current training pipelines. While recent method leverages a variety of emotion datasets, such as EMER~\cite{lian2023EMER}, MERR~\cite{cheng2024MERR}, and MER-Caption~\cite{lianaffectgpt25}, only a small fraction of these include human-annotated rationales. The majority of training relies on larger datasets without explanation labels, where reward signals focus solely on the correctness of predicted emotions, regardless of whether the generated rationale is emotionally coherent. As a result, models generate emotionally incoherent explanations without penalty, leading to a misalignment between predictions and rationales, as shown in Figure~\ref{fig:fig1}.

In this paper, we propose a novel framework for generating emotionally coherent explanations, even when trained on emotion datasets lacking human-annotated rationales. Our approach is based on the key assumption that LLMs possess sufficient linguistic and emotional knowledge to evaluate the consistency between a given emotion and a generated explanation. However, incorporating LLMs directly into the evaluation process can be computationally expensive and impractical in deployment settings. To address this, we distill the knowledge of the LLM into a lightweight model, referred to as the Emotional Rationale Verifier (ERV), which efficiently assesses the alignment between an explanation and the target emotion. Building on this verifier, we integrate ERV into a reward-driven training pipeline. In this setup, the model receives an explanation reward based on how well its generated explanation aligns with the target emotion as judged by ERV. To further improve reward fidelity, we decompose each explanation into individual sentences, filter out emotionally neutral ones (e.g., those describing background or appearance), and assign rewards based on the proportion of emotionally salient sentences.

Furthermore, to evaluate the emotional coherence, we introduce a set of novel metrics: Explanation Emotion Accuracy (EEA), which measures alignment between the explanation and the target emotion, Faithful Consistency Rate (FCR), requires all three components, explanation, predicted emotion, and target emotion, to agree, and Explanation–Prediction Consistency (EPC), which quantifies the consistency between the explanation and predicted emotion. These metrics provide deeper insight into the emotional coherence of generated explanations beyond standard classification accuracy.

Experimental results demonstrate that our approach preserves emotion classification performance while improving the consistency between generated explanations and predicted emotions, as measured by the three proposed evaluation metrics. Moreover, human evaluation shows that the model produces more coherent and emotionally aligned reasoning, thereby supporting more reliable and interpretable interactions in HCI applications.

Our key contributions are as follows:
\begin{itemize}
    \item \textbf{Emotionally consistent explanation framework:}  
    We propose a novel framework that improves the emotional coherence of explanations without modifying the model architecture or requiring labeled explanation data by leveraging CoT reasoning and reward-based learning.

    \item \textbf{Emotional Rationale Verifier (ERV):}  
    We introduce ERV, a lightweight verifier distilled from an LLM, which assesses the alignment between generated explanations and target emotions, and serves as a reward function to guide explanation generation.

    \item \textbf{New metrics for emotional coherence evaluation:}  
    We propose three novel metrics—EEA, EPC, and FCR—that enable fine-grained evaluation of emotional alignment and consistency in model-generated explanations.
\end{itemize}

\section{Related Works}
\subsection{Multimodal Emotion Recognition}
The task of Multimodal Emotion Recognition (MER) aims to understand human emotional states by analyzing diverse multimodal cues in expressive scenes. From audio and visual cues to text, body, and physiological signals, numerous studies~\cite{kossaifi2017afew,zadeh2018multimodal,poria2019meld,busso2008iemocap,perepelkina2018ramas,liu2022mafw,jiang2020dfew,livingstone2018RAVDESS,lee2019CAER,wang2022ferv39k} have proposed datasets consisting of emotion-related modalities and corresponding emotion labels.
With those datasets, traditional approaches have primarily relied on modality-specific encoders~\cite{zhang2017speech,zhang2021learning, zhao2021former,sun2023mae,chumachenko2024mma} for feature extraction and fusion strategies~\cite{narayan2025facexformer,sun2024hicmae}, achieving solid performance in emotion classification.

However, unlike object classification, emotion classification suffers from inherent ambiguity, as emotion labels alone often fail to fully capture an individual's state. To address the challenge of unreliable labeling caused by the limited expressivity of fixed label mappings, multimodal emotion descriptive dataset EMER~\cite{lian2023EMER} has been proposed.
This dataset requires models not only to predict emotion labels from visual, audio, and textual cues, but also to generate descriptive explanations of the emotional states.
Additionally, with the rise of Large Language Models (LLMs)~\cite{touvron2023llama2,team2024qwen2, grattafiori2024llama}, and their powerful natural language generation capabilities, recent instruction-tuned MLLM approaches~\cite{xing2024emo,cheng2024MERR,lianaffectgpt25,yang2025omni,hu2025feallm, zhao2025favchat} move beyond classification toward reasoning, enabling interpretable and context-aware emotion understanding.

Furthermore, motivated by the increasing use of rich emotional descriptions, recent studies~\cite{cheng2024MERR,lian2024open,yang2025omni} have explored leveraging the generative capabilities of LLMs to automatically expand emotion descriptive datasets. These approaches aim to enrich the diversity and expressivity of descriptive samples by generating additional explanations grounded in multimodal cues. Building on such augmented and description-rich emotion datasets, many works~\cite{cheng2024MERR,lian2024open} have investigated interpretable reasoning over video and audio, while architectural advancements~\cite{zhao2025humanomni,lianaffectgpt25} have further improved recognition accuracy. However, most existing methods rely on supervised fine-tuning (SFT), which limits generalization due to its dependence on the scale and diversity of annotated emotional descriptions.

\subsection{GRPO-Based RL with Verifiable Rewards}
Recent work has shown that RL post-training can substantially sharpen the CoT abilities of MLLMs. In particular, Group Relative Policy Optimization (GRPO)~\cite{shao2024deepseekmath} combined with Verifiable Reward~\cite{guo2025deepseek} achieves strong generalization with limited data and fixed model backbones, excelling in math and coding benchmarks.

Motivated by this success, GRPO has been adapted to a wide range of multimodal tasks, including emotion recognition~\cite{zhao2025r1}, general vision tasks~\cite{chen2025r1v}, object detection~\cite{liu2025visualrft,shen2025vlm,park2025dip}, and the video domain~\cite{li2025videochat, lee2025refocus,xing2025echoink}, often designing reward functions tailored to their specific task objectives. However, these approaches predominantly focus on optimizing the final answer, paying little attention to the quality and consistency of the intermediate reasoning steps, a degradation problem highlighted by ~\cite{wei2025gtr}.

To address this, recent studies~\cite{chen2025grpocare,yang2025humanomniv2} have introduced consistency-aware methodologies to improve the alignment between reasoning and the answer. Yet, those consistency problems have not yet been explored in the field of MER, where reasoning datasets are limited. This gap is significant, as MER applications require explanations that are not just plausible but emotionally coherent with the predicted affect.
We therefore introduce a novel reward applied with GRPO training that explicitly enforces coherence between the generated explanation and the target emotion, addressing the unique challenge of achieving consistency in the absence of human-annotated rationales.

\section{Method}
Our primary objective is to develop an MLLM capable of emotion understanding that not only accurately predicts emotion labels from face video and speech audio but also generates consistent and faithful natural language explanations aligned with these predictions. Building upon recent advancements demonstrating the effectiveness of RL-based training for improving both emotion recognition accuracy and reasoning in MLLM~\cite{zhao2025r1}, we integrate a novel Emotional Rationale Verifier (ERV) module into an established RL framework. This integration allows us to directly evaluate and enhance the emotional consistency of the generated explanations.
Specifically, we extend the HumanOmni model architecture~\cite{zhao2025humanomni} and GRPO~\cite{guo2025deepseek}, a reward-driven RL training strategy showing strong performance in various multimodal tasks~\cite{liu2025visualrft,chen2025r1v,zhao2025r1}. While HumanOmni effectively unifies multimodal features and GRPO enhances CoT reasoning, existing approaches often struggle with emotionally incongruent explanations despite accurate label predictions, as highlighted in our introduction. 

\begin{figure*}[t]
    \centering
    \includegraphics[width=1.0\textwidth]{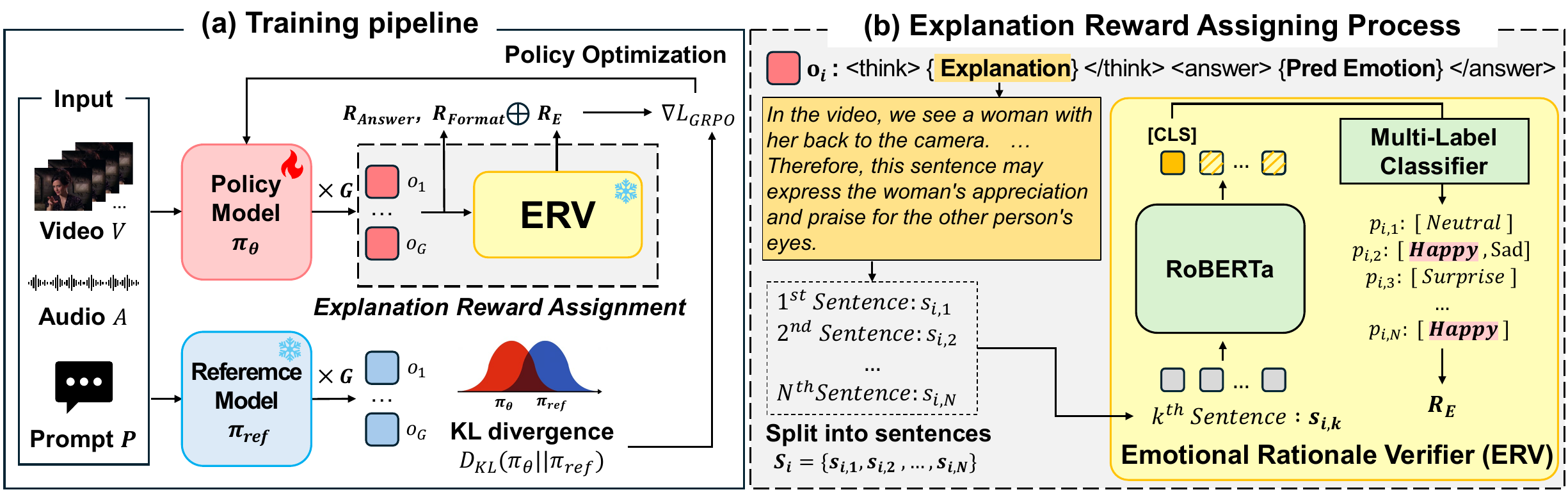}
    \caption{\textbf{(a):} The policy model $\pi_{\theta}$ generates $G$ responses $o_1$, ..., $o_G$, and the Emotional Rationale Verifier (ERV) assigns an explanation reward $R_E$ to each response. \textbf{(b):} For each response $o_i$, its explanation $E_i$ is extracted and evaluated to produce its reward $R_{i,E}$. The ground truth emotion used for evaluation in this scenario is `Happy'($e_{gt}=$ `Happy'). Collectively, $R_E = \{ R_{1,E}, R_{2,E}, \ldots, R_{G,E} \}$ denotes the set of explanation rewards for the $G$ responses.}
    \label{fig:fig2}
\end{figure*}

To precisely address this critical misalignment, our proposed ERV module assesses whether a generated explanation coherently supports the target emotion. Based on this assessment, we define a novel reward signal, the Explanation Reward ($R_E$), which encourages the MLLM to produce more faithful and emotion-aligned rationales. The overall training pipeline, detailing the ERV module and its integration into the GRPO framework, is visually represented in Figure~\ref{fig:fig2}.

\subsection{Model Architecture and Training Overview}
Our methodology leverages the HumanOmni architecture as its foundational MLLM. Given an input video $V$, audio $A$, and task prompt $P$, the model first processes the raw multimodal inputs using pretrained visual and audio encoders. These modality-specific features $X_v$ (visual) and $X_a$ (audio) are then projected into a shared embedding space and concatenated with the prompt embedding ($X_p$). The resulting comprehensive fused representation ($X_m$) is then fed into the Large Language Model (LLM), denoted as $\psi$, which autoregressively generates the output sequence ($o$):

{\small
\begin{equation}
\begin{aligned}
    X_m &= \operatorname{Concat}(X_v, X_a, X_p), & o = \psi(X_m)
\end{aligned}
\end{equation}
}

The training of this model follows a robust two-stage strategy: an initial Supervised Fine-Tuning (SFT)~\cite{zhang2024SFT} phase, followed by RL using GRPO. This progressive approach ensures the model first learns fundamental reasoning structures and then refines its explanation generation based on explicit reward signals.

\subsubsection{Initializing Emotion Reasoning via SFT}
The initial training phase involves SFT on a high-quality, albeit relatively small, dataset named EMER~\cite{lian2023EMER}. This dataset consists of $(V, A)$ pairs annotated with emotion labels and detailed explanatory descriptions. The primary objective of this phase is to enable the MLLM to grasp the basic structural relationships between multimodal inputs, emotion labels, and their textual explanations. It specifically helps the model learn to produce outputs in the required format $<$think$>$...$<$/think$>$$<$answer$>$...$<$/answer$>$. This SFT phase serves as a crucial initialization step for our policy model, denoted as $\pi_{\theta}$ (i.e., the trainable MLLM).
During SFT, the model is optimized to maximize the likelihood of generating the target output sequence given the input. The training objective is the negative log-likelihood loss:

{
\small
\begin{equation}
\mathcal{L}_{\text{SFT}} = - \mathbb{E}_{(x, y) \sim \mathcal{D}_{\text{SFT}}} \left[ \sum_{t=1}^{|y|} \log \pi_{\theta}(y_t \mid x, y_{<t}) \right]
\label{eq:sft_loss}
\end{equation}
}

Here, $(x, y)$ represents the input-output pair sampled from the SFT dataset $\mathcal{D}_{\text{SFT}}$, where $x$ represents the input triplet $(V, A, P)$ and $y$ is the corresponding explanation.

\subsubsection{Enhancing Explanation Coherence through Explanation-Guided Reinforcement Learning}
Following the SFT phase, we transition to RL using GRPO to enhance the model's reasoning capability and, critically, the consistency of its explanations. The SFT-trained model $\pi_{\theta}$ functions as the learnable policy, while a frozen duplicate, $\pi_{ref}$, acts as the reference model.
For each input $(V, A, P)$, both the policy model and the reference model generate G candidate output sequences, $\{o_1, o_2, \dots, o_G\}$. Each generated sequence is then evaluated by a comprehensive reward function, which we detail in the next section. The policy $\pi_{\theta}$ is subsequently optimized to maximize the expected reward $R$, while simultaneously being regularized to remain close to the behavior of $\pi_{ref}$. The optimization objective of GRPO is as follows:

{\small
\begin{equation}
\max_{\pi_\theta} \mathbb{E}_{o \sim \pi_\theta(x)} \left[ R(o)  - \beta \cdot \mathrm{KL}\left( \pi_\theta(o|x) \| \pi_{\text{ref}}(o|x) \right) \right]
\label{eq:grpo_loss}
\end{equation}
}

While the GRPO framework is highly effective for reward-driven optimization, it traditionally lacks an explicit mechanism to guarantee the consistency between the predicted emotion and its generated explanation. To overcome this inherent limitation and directly address the core challenge identified in our introduction, we introduce the Emotional Rationale Verifier ($ERV$) and its associated reward ($R_E$) as a novel and crucial signal within the GRPO training loop.

\subsection{Emotional Rationale Verifier (ERV): Design and Training}
We define an explanation-related reward signal based on how well the explanation in the output sequence $o$ reflects the target emotion. To assess the emotional content of the generated explanation \(E\), we employ an auxiliary module capable of classifying emotion labels from textual descriptions.

Although existing datasets such as EMER~\cite{lian2023EMER} and MERR-fine~\cite{cheng2024MERR} provide pairs of emotional descriptions and emotion labels, their limited sizes (332 and 4,487 samples, respectively) are insufficient for our setting. This is particularly problematic because the generated descriptions often contain not only emotional expressions but also contextual details about people, environments, and temporally evolving emotional states. In addition, the emotion label distribution is imbalanced.
As a result, existing datasets lack the diversity and coverage necessary to effectively train the ERV.

To address the limitations of existing datasets in terms of both size and class imbalance, we construct a pseudo-labeled dataset consisting of emotional text descriptions paired with emotion labels. As shown in Figure~\ref{fig:fig1}, while R1-Omni~\cite{zhao2025r1} does not always produce perfectly emotion-aligned outputs, its generated descriptions are often emotionally plausible. Leveraging this property, we generate textual explanations from the training sets of DFEW~\cite{jiang2020dfew} and MAFW~\cite{liu2022mafw}, and use a closed-source LLM (GPT-4.1) to assign up to two representative emotion labels to each description. This process yields a pseudo-labeled dataset of 20K emotional text–label pairs, which is used to train our ERV module. To further mitigate the class imbalance, we augment underrepresented categories, namely \textit{disgust}, \textit{fear}, \textit{contempt}, \textit{disappointment}, \textit{neutral}, and \textit{helplessness}, by generating additional emotional descriptions using GPT-4.1.

To train the ERV on the constructed dataset, we use the RoBERTa model trained on the GoEmotions dataset~\cite{demszky2020goemotions}\footnote{\url{https://huggingface.co/SamLowe/roberta-base-go_emotions}} as a backbone.  
We fine-tune this model to predict emotion labels from text descriptions using our 23K-sample dataset, which includes both human-annotated and GPT-generated pseudo-labeled examples.  
The model predicts the emotion label based on the output of the \texttt{[CLS]} token, which encodes the global semantics of the input description.  
These predictions are then used to compute the Explanation Reward for each generated explanation \(E_i\).

\subsection{Explanation Reward}
With the proposed ERV, we define an explanation reward \(R_E\), which is assigned in proportion to how well the generated explanation \(E\) reflects the GT emotion \(e_{\text{gt}}\). 
However, a long narrative often mixes various types of content such as scene background, appearance details, and affective cues, making it unreliable to evaluate the entire paragraph as a whole. Sentences that merely describe appearance or context can dilute the emotional signal and mislead ERV.

To address this, we design the explanation reward \(R_E\) with two key components.
We illustrate the reward computation using the \(i\)th output \(o_i\) from the policy model \(\pi_{\theta}\), as shown in Figure~\ref{fig:fig2}~(b).
\textbf{1) Sentence-level and multi-label verification.}  
For each generated sentence \(s_{i,k} \in E_i\), ERV predicts a (possibly multi-label) emotion set \(p_{i,k}\).  
This sentence-level prediction prevents non-emotive sentences from dominating the overall judgment.  
Furthermore, since a single sentence may convey multiple emotions, we apply a threshold \(\tau\) to extract up to two emotion labels per sentence from the classifier.
\textbf{2) Neutral-sentence filtering.}  
If ERV classifies \(s_{i,k}\) as \textsc{neutral}, the sentence is excluded from the computation of \(R_{i,E}\), based on the intuition that surface-level descriptions such as listing physical traits or environmental details provide little to no emotional signal and thus do not meaningfully contribute to emotion understanding.

Given the set of split sentences \(S_i\), each sentence is evaluated by ERV.  
We count the number of sentences whose predicted emotions \(p_{i,k}\) match the GT emotion \(e_{\text{gt}}\) as \(c_i\), and denote the number of sentences classified as \textsc{neutral} as \(N_{i,\text{Neu}}\).  
The explanation reward \(R_{i,E}\) is then computed as:
{
\small
\begin{equation}
R_{i,E} =
\begin{cases}
\displaystyle \frac{c_i}{N_i - N_{i,\text{Neu}}}, & \text{if } e_{\text{gt}} \ne \textsc{neutral}, \\[10pt]
\displaystyle \frac{c_i}{N_i}, & \text{if } e_{\text{gt}} = \textsc{neutral}.
\end{cases}
\label{eq:Eq4}
\end{equation}
}
\paragraph{Total reward.}
In addition to the \(R_E\), we incorporate two auxiliary rewards \textit{format reward} \(R_{\text{format}}\) and \textit{answer reward} \(R_{\text{answer}}\) to compute the total reward \(R\).  
\(R_{\text{format}}\), following prior work~\cite{guo2025deepseek}, assigns 1 if the model output adheres to the required format (i.e., includes both \texttt{<think>} and \texttt{<answer>} tags), and 0 otherwise.  
\(R_{\text{answer}}\) assigns 1 if the predicted emotion within the \texttt{<answer>} matches \(e_{\text{gt}}\), and 0 otherwise.
The total reward used in training is computed as:

{
\small
\begin{equation}
R = R_E + R_{\text{format}} + R_{\text{answer}}.
\end{equation}
}
Incorporating $R_E$ during GRPO training significantly enhances consistency between the model's reasoning and the target emotion.

\begin{table*}[t]
\centering
\renewcommand{\arraystretch}{1.2}
\resizebox{\textwidth}{!}{%
\begin{tabular}{
    c !{\vrule width 1pt}
    c c c c c !{\vrule width 1pt}
    c c c c c
}
\toprule
\multirow{2.5}{*}{\textbf{Model}} 
  & \multicolumn{5}{c!{\vrule width 1pt}}{\textbf{MAFW}} 
  & \multicolumn{5}{c}{\textbf{DFEW}} \\
\cmidrule(lr){2-6} \cmidrule(lr){7-11}
  & \textbf{EEA (\%)}
  & \textbf{FCR (\%)}
  & \textbf{EPC (\%)}
  & \textbf{WAR (\%)}
  & \textbf{UAR (\%)}
  & \textbf{EEA (\%)}
  & \textbf{FCR (\%)}
  & \textbf{EPC (\%)}
  & \textbf{WAR (\%)}
  & \textbf{UAR (\%)} \\
\hline

\rowcolor[HTML]{EFEFEF}
\multicolumn{11}{c}{\textbf{\centering SFT-based Emotion MLLMs}} \\

Human-Omni~\cite{zhao2025humanomni}   & –    & –    & -         & 20.18 & 13.52     & –      & –      & -     & 22.64 & 19.44  \\
EMER-SFT       & 33.06 & 29.67 & \textbf{70.66}   & 38.95 & 25.21 & 35.14 & 32.63 & 64.30 & 38.93 & 30.86 \\
EMER-SFT (7B)    & 44.26 & 38.58 & 67.38   & 50.22 & 33.48 & 44.86 & 42.21 & 57.41 & 48.33 & 39.16 \\
\textit{Emotion-LLaMA (7B)~\cite{cheng2024MERR}} & - & - & - & - & - & 41.15 & 32.56 & 57.69 & 48.71 & 38.54  \\  
\textit{Emotion-LLaMA (7B) Fine-Tuned} & - & - & - & - & - & - & - & - & 77.06 & 64.21  \\ 
\rowcolor[HTML]{EFEFEF}
\multicolumn{11}{c}{\textbf{\centering RL-based Emotion MLLMs}} \\

R1-Omni~\cite{zhao2025r1}        & 37.87 & 31.80 & 48.09   & \textbf{57.68} & \textbf{40.04} & 43.69 & 37.99 & 51.91 & 65.83 & \textbf{56.27} \\
Baseline       & 32.41 & 27.27 & 40.98   & 56.83 & 38.66 & 41.66 & 36.39 & 48.67 & 66.17 & 55.33  \\
\textit{Baseline (7B)} & 44.97 & 40.22 & 55.19 & \underline{65.36} & \underline{47.27} & 55.39 & 52.01 & 60.70 & \underline{76.67} & \underline{69.69} \\  
\rowcolor[HTML]{EFEFEF}
\multicolumn{11}{c}{\textbf{\centering Ours}} \\

\textbf{Ours}  & \textbf{48.93} & \textbf{43.57} & 68.58 & 57.49 & 39.12 & \textbf{59.45} & \textbf{53.75} & \textbf{73.06} & \textbf{67.30} & 56.26 \\
\textbf{Ours (7B)} & \underline{54.70} & \underline{50.98} & \underline{73.06} & 65.19 & 47.01 & \underline{65.50} & \underline{62.06} & \underline{73.53} & 75.81 & 68.88 \\  
\bottomrule
\end{tabular}%
}
\caption{Evaluation of models on proposed metrics: Explanation Emotion Accuracy (EEA), Faithful Consistency Rate (FCR), Explanation–Prediction Consistency (EPC), and classification accuracy (WAR, UAR) on MAFW and DFEW datasets. Underline indicates the best score based on the 7B-sized model, while bold indicates the best score based on the 0.5B-sized model.}
\label{tab:tab1}
\end{table*}

\section{Experimental Setup}
\subsection{Dataset}
\noindent\textbf{EMER}~\cite{lian2023EMER} is a human-annotated emotional video dataset with 332 samples labeled with one of five emotions (\textit{angry}, \textit{sad}, \textit{surprise}, \textit{worried}, \textit{happy}) and corresponding descriptions. It is used to train the model to generate explanations in a reasoning format.

\noindent\textbf{MERR}~\cite{cheng2024MERR} provides 4,487 fine-grained pseudo-labeled video-caption pairs, generated via facial and audio analysis with LLM-guided annotation. We use this data to supervise our 
ERV, leveraging its diverse and component-level emotion cues.

\noindent\textbf{DFEW}~\cite{jiang2020dfew} is a large-scale facial expression video dataset with 9,362 training and 2,342 test samples, labeled with 7 basic emotions: \textit{angry}, \textit{disgust}, \textit{surprise}, \textit{happy}, \textit{sad}, \textit{neutral}, and \textit{fear} (official \texttt{5th split}). We use DFEW in two ways: (1) generating textual explanations from training videos and assigning pseudo-labels via GPT-4.1 to augment ERV training, and (2) applying our reward-based RL scheme.

\noindent\textbf{MAFW}~\cite{liu2022mafw} is a multimodal emotion dataset with 7 basic and 4 compound categories (\textit{contempt}, \textit{helplessness}, \textit{anxiety}, \textit{disappointment}), comprising 7,341 training and 1,831 test samples (\texttt{set 5 split}).
We apply the same pipeline as DFEW.

\subsection{Evaluation Metrics}
\noindent\textbf{Emotion Recognition Accuracy} Following prior works~\cite{zhao2025r1}, we employ Unweighted Average Recall (UAR) and Weighted Average Recall (WAR) to measure the model’s emotion classification performance.

\noindent\textbf{Evaluating Emotional Coherence} We evaluate the emotional coherence of generated explanations using three metrics: Explanation Emotion Accuracy (EEA), Explanation–Prediction Consistency (EPC), and Faithful Consistency Rate (FCR). These metrics capture different aspects of alignment among the explanation, predicted emotion, and ground-truth label. GPT-4.1-mini is used to recognize emotion in explanation. Figure~\ref{fig:fig3} shows the prompt used in evaluation.

\begin{figure}[t]
\centering
\fbox{%
\begin{minipage}{0.95\linewidth}
\small
\texttt{Read the reasoning content and respond with the appropriate emotion in\\ \{Emotion List\}.}\\
\texttt{Reply with only the emotion word.}\\
\texttt{Reason: \{Explanation\} Answer Emotion:}
\end{minipage}
}
\caption{Prompt used to evaluate the emotion conveyed in the generated explanation. \{Emotion List\} is a permutation of the GT emotion label according to the evaluation dataset. \{Explanation\} is $E_i$ from output $o_i$.}
\label{fig:fig3}
\end{figure}

\subsection{Implementation Details}
We employ the HumanOmni 0.5B model~\cite{zhao2025humanomni} as our backbone and RoBERTA-100M (trained on GoEmotions) for the ERV. Our training pipeline consists of three components: SFT, ERV, and GRPO. The main model is first fine-tuned on the EMER dataset for 5 epochs with a $2\mathrm{e}^{-5}$ learning rate, a cosine scheduler, a 0.03 warmup ratio, and a batch size of 32. 
The ERV module is trained separately for 10 epochs with a batch size of 64 and a $3\mathrm{e}^{-5}$ learning rate. Those are trained on 8 NVIDIA RTX 3090 GPUs.
For GRPO training, we use a batch size of 16 with 2 gradient accumulation steps, generating $G=16$ samples per input, and apply a KL divergence penalty of $0.04$. For inference, we use a temperature of 0.3. For the 7B model, we trained with $G=4$ instead. This stage is conducted on 8 NVIDIA A100 GPUs. Appendix~\ref{appendix: training_details} shows more training details.

\section{Experimental Results}
\subsection{Main Results}

To verify the effectiveness of our proposed Explanation reward in the quality of generated explanations, we evaluated on three proposed metrics: EEA, EPC, and FCR, with standard recognition accuracy WAR and UAR. 
Our model is compared with three models trained in different ways using the same multimodal emotion description dataset and HumanOmni architecture. (1) \textbf{EMER-SFT}: A supervised fine-tuned model trained on the EMER dataset for emotion reasoning, available in both 0.5B and 7B. (2) \textbf{R1-Omni}: A model trained with the GRPO using answer and format rewards. (3) \textbf{Baseline}: Since R1-Omni lacks training details and relies on an in-house dataset, we re-trained EMER-SFT with GRPO on the answer and format rewards using the same training dataset and hyperparameter settings as our method for a fair comparison.

As shown in Table~\ref{tab:tab1}, both EMER-SFT and R1-Omni exhibit low FCR, achieving only 29.67\% and 31.80\%, respectively. This indicates that although R1-Omni significantly improves emotion recognition accuracy on WAR, from 38.95\% to 57.68\% compared to EMER-SFT, 44.87\% of its predictions (almost half) are not supported by emotionally aligned explanations. Furthermore, its ability to independently produce explanations aligned with the target emotions (EEA) improves only marginally from 33.06\% to 37.87\%.
In contrast, our model trained with an Explanation reward not only achieves comparable recognition performance to R1-Omni but also demonstrates substantial improvements in both FCR and EEA. FCR increases from 29.67\% to 43.57\% and also EEA increases from 33.06\% to 48.93\%. A similar tendency is observed on the DFEW, where FCR improves from 32.63\% to 53.75\% and EEA improves from 35.14\% to 59.45\%.

We also compared our model with EMER-SFT (7B) and Emotion-LLaMA (7B), both of which use 7B-sized backbones but have different architectures. In the evaluation on the DFEW dataset, both models exhibited low performance, achieving EEA scores of 44.86\% and 41.15\%, and FCR scores of 42.21\% and 32.56\%, respectively. These results indicate that SFT training on existing multimodal video–reasoning pair datasets yields unreliable emotion reasoning performance regardless of the model architecture. In contrast, despite using a smaller 0.5B LLM size, our model outperforms both baselines in terms of FCR and EEA, while also achieving noticeably better emotion recognition accuracy. When applied to 7B-scale models, our method shows comparable recognition accuracy to Emotion-LLaMA fine-tuned solely for emotion label prediction, while achieving 62.06\% FCR and 65.50\% EEA.

About EPC, EMER-SFT shows significant performance compared to RL-based trained models. Namely, the final predicted answer is derived based on the generated explanations. However, when trained with GRPO on answer and format reward, this consistency deteriorates, as the model tends to focus solely on producing the final answer, leading to a degradation in EPC. In contrast, our proposed Explanation Reward enables the model to maintain this consistency, achieving performance that is comparable to or even surpasses that of EMER-SFT (70.66\% vs. 68.58\% in MAFW and 64.30\% vs. 73.06\% in DFEW). This indicates that our Explanation Reward effectively guides the model to generate explanations and predictions in a coherent and aligned manner.

\begin{table}[t]
    \centering
    \small     
    \setlength{\tabcolsep}{2pt} 
    
    \begin{tabularx}{\linewidth}{l *{5}{>{\centering\arraybackslash}X}} 
    \toprule
    \makecell[b]{\textbf{Setting}} & 
    \makecell[b]{\textbf{EEA}} & 
    \makecell[b]{\textbf{FCR}} &
    \makecell[b]{\textbf{EPC}} &
    \makecell[b]{\textbf{WAR}} &
    \makecell[b]{\textbf{UAR}} \\
    \midrule
    \hspace{2mm} Baseline & 32.41 & 27.27 & 40.98 & 56.83 & 38.66 \\
    \midrule
    \multicolumn{6}{l}{\textbf{ERV}} \\
    \midrule
    \hspace{2mm} \makecell[l]{W/o Data Aug.} & 46.88 & 42.62 & 65.46 & 56.78 & 38.60 \\ 
    \midrule
    \multicolumn{6}{l}{\textbf{Explanation Reward ($R_E$)}} \\ 
    \midrule
    \hspace{2mm} W/o Neutral Filt. & 46.28 & 42.13 & 64.70 & 57.16 & 38.67 \\ 
    \hspace{2mm} \makecell[l]{W/o Sent. Verif.} & 47.54 & 43.17 & 68.36 & 56.78 & 38.66 \\ 
    \midrule
    \hspace{2mm} Ours & \textbf{48.93} & \textbf{43.57} & \textbf{68.58} & \textbf{57.49} & \textbf{39.12} \\
    \bottomrule
    \end{tabularx}
    \caption{Ablation study of the ERV and $R_E$ on MAFW. All values are reported in percentage (\%).}
    \label{tab:tab2}
\end{table}

\subsection{Ablation Study}
\subsubsection{Effectiveness of Emotional Rationale Verifier}
To verify our proposed training strategy on the ERV module, we removed the data augmentation stage. (1) Baseline, without the ERV module, can't reward on the generated explanation. It only focuses on the format and answer. 
(2) Without a data augmentation strategy, training the ERV solely on the EMER and MERR-fine datasets boosts EEA from 32.41\% to 46.88\% and FCR from 27.27\% to 42.62\%, while keeping WAR and UAR unchanged. (3)  Adding the augmented data to ERV further raises EEA + 1.26 pp, FCR + 1.31 pp, EPC + 2.90 pp, and overall recognition accuracy.
These results suggest that the ERV’s judgment capability is constrained by the size of its training data. When additional multimodal-emotion descriptions are unavailable, text-level data augmentation noticeably enhances the model’s multimodal reasoning ability.
\subsubsection{Effectiveness of Explanation Reward}
To validate the effectiveness of our proposed Explanation Reward, we compare it against a variant trained without our two key components: (1) Neutral Filtering and (2) Sentence-Level Verification. In this variant, the reward calculation does not filter out non-emotional sentences (as described in Equation~\ref{eq:Eq4} for the neutral case), and the ERV evaluates the emotional coherence of the entire explanation holistically. We observed a performance drop when each of these components was removed individually. Specifically neutral filtering, which excludes emotionally irrelevant sentences, was particularly effective in guiding the model to generate emotion-focused explanations.

\begin{table}[t!]
\centering
\small
\begin{tabular}{lcc}
\toprule
\textbf{Evaluation} & \textbf{Ours} & \textbf{R1-Omni} \\
\midrule
Win Rate &  53.6\%  &  46.4\%  \\
Assessment Score &  3.52  &  3.42  \\
\bottomrule
\end{tabular}
\caption{Human evaluation comparison between our model and R1-Omni}
\label{tab:tab3}
\end{table}

\begin{table}[t!]
\centering
\small
\begin{tabular}{l l c c c}
\toprule
\textbf{Closed-source LLM} & \textbf{Model} & \textbf{EEA} & \textbf{FCR} & \textbf{EPC} \\
\midrule
\multirow{2}{*}{GPT-4.1-mini} 
& Baseline & 32.41 & 27.27 & 40.98 \\
& Ours & 48.93 & 43.57 & 68.58 \\
\midrule
\multirow{2}{*}{Gemini-Flash-2.5} 
& Baseline & 32.24 & 27.76 & 41.80 \\
& Ours & 48.52 & 42.73 & 65.96 \\
\bottomrule
\end{tabular}
\caption{Performance of Baseline and Ours (EEA, FCR, EPC) using different closed-source LLMs. All values are reported in percentage (\%).}
\label{tab:tab4}
\end{table}

\subsection{Human Evaluation about Explanation Quality}
Through the main results and ablation studies, we confirmed that explanations from our model better align with both target and predicted emotions.
However, their quality in terms of how well they explain the emotions in the videos had not been directly evaluated.
To address this, we conducted a human study comparing our model’s explanations to those from R1-Omni. Specifically, we randomly selected 14 samples from each model (total 28), where the generated emotion matched the ground-truth.
Twenty human raters evaluated each explanation on a 1–5 scale based on how well it explained the corresponding video’s emotion. For each sample, we recorded which model received the higher score and computed the average score across all samples.
As shown in Table~\ref{tab:tab3}, our model achieved a 53.6\% win rate and a higher average score (3.52 vs. 3.42), indicating that it produces more informative and emotionally grounded explanations than R1-Omni. (Survey details are provided in Appendix~\ref{appendix:human_survey}.)

\subsection{Robustness of the Proposed Reasoning Evaluation Metrics}
Since the evaluation of EEA, FCR, and EPC was conducted using a single closed-source LLM, we further validated the robustness of the proposed metrics by repeating the evaluation with a different model family, \textit{Gemini-2.5-Flash}~\cite{comanici2025gemini}.
As shown in Table~\ref{tab:tab4}, the evaluation on MAFW reveals a consistent trend, with both the Baseline and our model exhibiting improved reasoning quality. This demonstrates that our metrics are independent of the specific LLM evaluator and consistently capture the alignment between reasoning and emotion across model families. Moreover, a human survey confirmed a 92.9\% agreement between participants' judgments on the conveyed emotion and those of GPT-4.1-mini. (See Appendix~\ref{appendix:human_survey}.)

\section{Conclusion}
We proposed training an MLLM for the MER task utilizing our proposed ERV and Explanation Reward, aiming to enhance the coherence between emotion recognition and explanation. Compared to prior approaches, where coherence between predictions and explanations among correctly predicted samples remained around half (55.13\% on MAFW and 42.92\% on DFEW), our method significantly improves this alignment, achieving 75.79\% and 59.77\%.Furthermore, human evaluations confirmed that our approach produces more emotionally and contextually grounded emotion explanations for video data. As a result, our work enables emotionally coherent and trustworthy explanations alongside accurate emotion recognition, representing a significant advancement toward emotionally intelligent human–computer interaction.

\section*{Acknowledgments}
This work was partly supported by two funds: Institute of Information \& communications Technology Planning \& Evaluation (IITP) grant funded by the Korea government (MSIT) (No. RS-2025-25442384, QuestioningAI: Development of Active Reasoning Cooperative Multimodal Agent Technology), and the National Research Foundation of Korea (NRF) grant funded by the Korea government (MSIT) (No. NRF-2022R1A2C2005529). Additionally, this work was supported by the National Supercomputing Center with supercomputing resources including technical support (KSC-2025-CRE-0090).

{
\fontsize{9.0pt}{10.0pt}\selectfont
\bibliography{aaai2026}
}

\clearpage
\appendix
\section{Human Evaluation on Explanation Quality}  
\label{appendix:human_survey}
Following the human evaluation described in Section 5.3, we detail here the two main purposes: (1) to assess the alignment between human judgments and a closed-source LLM (e.g., GPT-4.1-mini) regarding the representative emotion in each text explanation, and (2) to compare how well the explanations from different models capture the emotional content expressed in the video.
\subsection{Survey Questions}
We provided three questions for each sample, as show in Figure~\ref{fig:survey_prompt} and ~\ref{fig:Appendix_survey}.

\begin{table*}[t!]
\centering
\scriptsize
\begin{tabular}{l c c c c c c c c c c c}
\toprule
Dataset & Angry & Happy & Sad & Surprise & Anxiety & Disgust & Fear & Neutral & Contempt & Disappointment & Helplessness  \\
\midrule
EMER        &   115   &   92   &  43   &     23     &   59      &    -     &   -   &     -    &    -    &     -        &   -      \\
MERR-Fine   &   875   &   476   &  1,091   &   213    &  -   &       -     &   85   &     1,352    &     40     &     -          &      -     \\
+ DFEW (Train)        &   2,834   &   1,534  &   1,566  &   1,558       &    2,769     &     146    &  1,195    &   515      &    38      &      582       &    737       \\
+ MAFW (Train)   & 2,299    &   1,046   &   1,397   &   1,396  &   1,939       &     123    &     780    &   380   &  38    &    653      &      645    \\
+ Complemented   & -    &   -   &   -   &  -  &   -       &    600    &    -    &   500  &  800   &   200      &     200   \\
\bottomrule
\end{tabular}
\caption{Statistics of the existing and augmented datasets used for training ERV. Datasets with a $+$ in their names refer to augmented datasets consisting of emotional text description–emotion label pairs.}
\label{tab:pseudo_datasets}
\end{table*}

\textbf{Question 1} asks for the representative emotion conveyed in the generated explanation. When tested on the output from our proposed model, the predictions showed 92.9\% agreement with GPT-4.1-mini's predictions. When considering the top 2 candidate emotions, the agreement increased to 96.4\%.

\textbf{Questions 2 and 3} evaluate how well the explanations reflect the emotion expressed in the video, not just by labeling an emotion. In other words, the focus is on how effectively the explanation interprets and describes the emotional signals embedded in the video. Specifically, each question is associated with one of the comparison models. These questions are rated on a 1-5 scale.

\begin{itemize}
    \item 5 = Reflects the emotion in the video very well
    \item 4 = Reflects the emotion in the video well
    \item 3 = Somewhat reflects the emotion in the video
    \item 2 = Poorly reflects the emotion in the video
    \item 1 = Does not reflect the emotion in the video at all
\end{itemize}

\begin{figure}[h]
\fbox{%
\begin{minipage}{0.95\linewidth}
\small
\texttt{Q1. \{Emotion Explanation\} What emotion does the explanation represent?}\\
\\
\texttt{Q2\&3.(Input Video) \{Emotion Explanation\} How well does this explanation reflect the emotion in the video?}
\end{minipage}
}
\caption{Two questions used to evaluate the emotion conveyed in the generated explanation. \{Emotion Explanation\} can be a generated output from each R1-Omni and our model. (Input Video) means video is accompanied by the question.}
\label{fig:survey_prompt}
\end{figure}

\begin{figure}[H]
    \centering
    \includegraphics[width=0.48\textwidth]{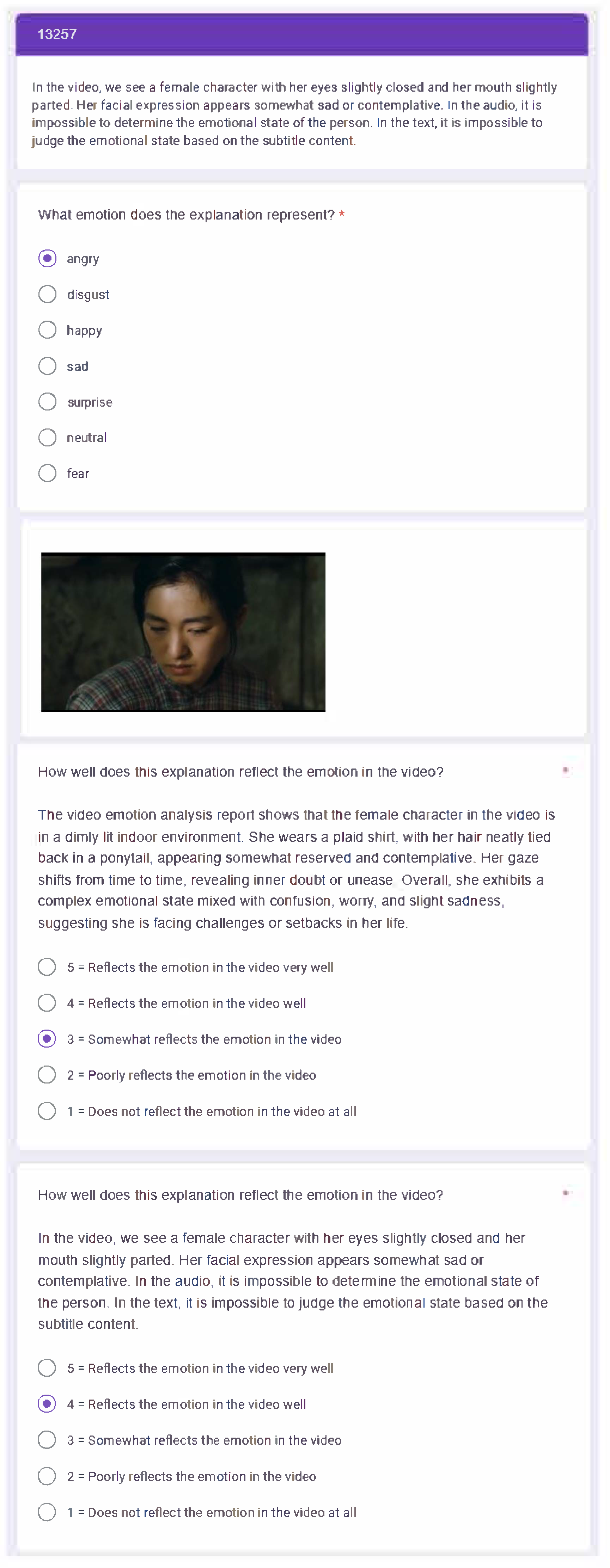}
    \caption{The actual survey platform provided to participants in the human study.}
    \label{fig:Appendix_survey}
\end{figure}

\subsection{Statistics on the Human Assessment Scores}
Table~\ref{tab:assessment} presents the mean and standard deviation of the human assessment scores. The evaluated explanations had already been identified by the closed-source LLM as expressing the target emotion. In other words, these samples were included in our proposed EEA metric. Overall, these samples demonstrated performance above a score of 3, indicating that the explanations appropriately describe the emotions conveyed in the videos.

\begin{table}[h]
\centering
\begin{tabular}{l c}
\toprule
Model & Assessment Score \\
\midrule
R1-Omni & 3.42 ± 0.34 \\
Ours & 3.52 ± 0.41 \\
\bottomrule
\end{tabular}
\caption{Mean and standard deviation of human assessment scores for emotional explanation quality across models.}
\label{tab:assessment}
\end{table}

\section{Augmenting Dataset for Training ERV}

\subsection{Details about Augmenting Emotional Text Descriptions}
To address the limited dataset sizes of EMER and MERR-Fine, we augmented emotional text descriptions paired with emotion labels. Using the MAFW and DFEW training sets along with the R1-Omni model, we generated emotional descriptions for each video. However, as shown in Figure~\ref{fig:fig1}, the R1-Omni model fails to generate explanations that align with the ground-truth emotions. However, it is still capable of producing emotionally rich textual descriptions in and of themselves, even if they do not match the annotated labels.

\begin{figure}[h]
\centering
\fbox{%
\begin{minipage}{0.95\linewidth}
\small
\texttt{Read the reasoning content and respond with the appropriate emotion in \{Emotion List\}.  Primarily output a single dominant emotion. If necessary, output up to 2 dominant emotions, separated by commas.}\\
\texttt{Reply with only the emotion word.}\\
\\
\texttt{Reason: \{Explanation\} Answer Emotion:}
\end{minipage}
}
\caption{Prompt used to generate the emotion conveyed in the generated explanation. \{Emotion List\} is a permutation of the GT emotion label from MAFW, and \{Explanation\} refers to the explanation generated by R1-Omni.}
\label{fig:generate_pseudo}
\end{figure}

Leveraging this property, we generated pseudo-emotional text descriptions from the DFEW and MAFW datasets. We then assigned emotion labels to these descriptions using GPT-4.1 with the prompt shown in Figure~\ref{fig:generate_pseudo}. This prompt is designed to output multi-label annotations from the given descriptions. The dataset statistics of MAFW and DFEW in Table~\ref{tab:pseudo_datasets} reflect overlapping multi-label annotations. As shown in the table, the resulting dataset exhibits class imbalance, with certain emotions such as \textit{disgust}, \textit{contempt}, \textit{neutral}, \textit{disappointment}, and \textit{helplessness} being underrepresented.

To mitigate this issue, we generated additional emotional text descriptions specifically for these underrepresented classes using GPT-4.1-mini, guided by the prompt illustrated in Figure~\ref{fig:Pseudo_fine_generating_prompt}. Two-shot examples, randomly selected from the pseudo dataset for each specific emotion, are provided to the LLM to generate the "Complemented" dataset. This dataset complements class-imbalanced emotion descriptions.

\begin{figure}[h]
\fbox{%
\begin{minipage}{0.95\linewidth}
\small

\texttt{You are an expert in audiovisual analysis with a talent for describing emotions.\\
Goal: Carefully interpret the video’s visual, auditory, and contextual elements, then portray the Target Emotion in vivid language.\\
\\
Example 1 (Target Emotion: \{Specific Emotion\}):\\
\{Example Description 1\}\\
\\
Example 2 (Target Emotion: \{Specific Emotion\}):\\
\{Example Description 2\}\\
\\
Now, based on the examples and the guidelines above, generate a new description that effectively conveys the following target emotion:\\
\\
Target Emotion: \{Specific Emotion\}
}
\end{minipage}
}
\caption{Prompt used to generate emotion-oriented explanations from audiovisual content. \{Emotion Explanation\} refers to model outputs (e.g., R1-Omni or our model), and the (Input Video) refers to the input video used in the task.}
\label{fig:Pseudo_fine_generating_prompt}
\end{figure}

\section{Metric Definitions}
\label{appendix:metrics}
We define three metrics to evaluate the emotional coherence between generated explanations, emotion label predictions, and ground-truth emotions.

Let $e_i$ denote the emotion predicted from the model's generated reasoning, $\hat{y}_i$ denote the model’s predicted emotion label, and $y_i$ denote the ground-truth emotion for the $i$-th sample. Then, the metrics are defined as follows:

\begin{equation}
\begin{aligned}
\mathrm{EEA} &= \frac{1}{N} \sum_{i=1}^{N} \mathbb{1}[e_i = y_i], \\
\mathrm{FCR} &= \frac{1}{N} \sum_{i=1}^{N} \mathbb{1}[e_i = y_i \land \hat{y}_i = y_i], \\
\mathrm{EPC} &= \frac{1}{N} \sum_{i=1}^{N} \mathbb{1}[e_i = \hat{y}_i].
\end{aligned}
\end{equation}

Here, $\mathbb{1}[\cdot]$ represents the indicator function that equals 1 if the condition within the brackets holds, and 0 otherwise. 

\textbf{EEA} (Explanation Emotion Accuracy) measures how well the emotion expressed in the generated explanation is aligned with the ground-truth emotion.

\textbf{FCR} (Faithful Consistency Rate) measures how well both the generated explanation and the predicted answer are aligned with the ground-truth emotion.  

\textbf{EPC} (Explanation–Prediction Consistency) measures how consistently the model’s predicted answer aligns with the generated explanation.

\section{Analysis of the Effectiveness of Explanation Reward}
We analyzed the changes in sample distribution by comparing the Baseline with our model trained with the Explanation Reward. In addition to the EEA metric, which evaluates whether the explanation correctly reflects the emotion, and the FCR metric, which measures whether both the explanation and answer are correct, we also investigated how much the proposed issue of predicting only the answer correctly has been alleviated. The proportion of such cases decreased from 29.56\% to 13.99\%, and we observed that this shift led to an FCR of 43.50\% due to improved explanation accuracy.
\begin{table}[h]
\centering
\begin{tabular}{c c|c c}
\toprule
Explanation & Answer & Baseline & Ours \\
\midrule
\checkmark    & \checkmark    & 27.27\%   & 43.50\% \\
\checkmark    & \ding{55}     & 5.14\%    & 5.36\% \\
\ding{55}     & \checkmark    & 29.56\%   & 13.99\% \\
\ding{55}     & \ding{55}     & 38.03\%   & 37.16\% \\
\midrule
\multicolumn{2}{c|}{EPC}       & 40.98\%   & 68.58\% \\
\bottomrule
\end{tabular}
\caption{Effect of Explanation Reward on Sample Distribution}
\label{tab:Quarter}
\end{table}


\section{Training Details}
\label{appendix: training_details}
We provide a more detailed description of the training configurations used in the SFT and GRPO training as shown in Table~\ref{tab:training_details}. We evaluated Emotion-LLaMA using the provided DFEW features and the checkpoint specialized for the emotion reasoning task. In the zero-shot evaluation on DFEW reported by Emotion-LLaMA, the model generated reasoning outputs for only about 40\% of the test samples, so the reasoning metric could not be clearly measured.

\begin{table}[h!]
\centering
\tiny
\begin{tabular}{lcccc}
\toprule
\textbf{configurations} & \textbf{SFT} & \textbf{SFT (7B)} & \textbf{GRPO} & \textbf{GRPO (7B)}\\
\midrule
Number of frames   & & \makecell[c]{8 uniformly\\sampled frames}& & \\
Image Resolution   &   224 × 224     & 384 × 384 & 224 x 224 & 384 × 384 \\
Lr scheduler       &   cosine     & cosine &  cosine-decay & cosine-decay\\
Optimizer          &  & \makecell[c]{AdamW\\($\beta_1=0.9$,\\$\beta_2=0.999$)} &\\
DeepSpeed    & \multicolumn{3}{c}{ZeRO-3} \\
Training precision & \multicolumn{3}{c}{bf16} \\
Total epochs       & 5            & 10         & 2 & 2 \\
Warmup ratio      & 0.03   & 0.03      & - & - \\
Learning rate  & 2e-5        & 2e-5      & 1e-6 & 1e-6 \\
Batch size  & 32      & 32      & 16  & 16 \\
\shortstack{Gradient\\accumulation}  &   1  &   1     & 2  & 2 \\
Trainable params   & \makecell[c]{language model,\\ vision tower,\\visual projector,\\audio projector,\\ text gate module} & \makecell[c]{language model,\\ vision tower,\\visual projector,\\audio projector,\\ text gate module} & \makecell[c]{language model,\\visual projector,\\audio projector,\\ text encoder} & \makecell[c]{language model,\\visual projector,\\audio projector,\\ text encoder} \\
Vision Encoder &
\makecell[c]{siglip-base\\(patch16-224)} &
\makecell[c]{siglip-so400m\\(patch14-384)} &
\makecell[c]{siglip-base\\(patch16-224)} &
\makecell[c]{siglip-so400m\\(patch14-384)}\\
Audio Encoder & \multicolumn{3}{c}{whisper-large-v3}  \\
\bottomrule
\end{tabular}
\caption{Training specific configurations}
\label{tab:training_details}
\end{table}

\clearpage

\section{Computing Details of Explanation Reward}
Algorithm~\ref{alg:explanation_reward}  details the procedure for computing the Explanation Reward $R_E$ for the output $O$ generated by $\pi_{\theta}$ over $G$ generations, as introduced in Figure~\ref{fig:fig2} and Equation~\ref{eq:Eq4}.

\begin{algorithm}[h!]
\caption{Pseudocode for Explanation Reward}
\label{alg:explanation_reward}

\textbf{Input}: \\
\hspace{0.8em} $G$: Number of generations for GRPO training \\
\hspace{0.8em} $O = \{o_i\}_{i=1}^{N_{gen}}$: Output sequences from $\pi_{\theta}$  \\
\hspace{0.8em} $ERV$: ERV module returning logits \\
\hspace{0.8em} $e_{gt}$: Ground-truth emotion label \\

\textbf{Parameters}: \\
\hspace{0.8em} $\tau = 0.5$: Probability threshold for emotion selection \\
\hspace{0.8em} $k = 2$: Maximum number of emotions selected per sentence \\

\textbf{Intermediate Variables}: \\
\hspace{0.8em} $E_i$: Extracted explanation from $o_i$ between \texttt{<think>} and \texttt{</think>} \\
\hspace{0.8em} $S_i = \{s_{i,1}, s_{i,2}, \dots, s_{i,l}\}$: Set of sentences split from $E_i$ \\
\hspace{0.8em} $c_i$: Number of sentences in $S_i$ consistent with $e_{gt}$ \\
\hspace{0.8em} $N_{i,neu}$: Number of neutral sentences in $S_i$\\
\hspace{0.8em} $EL_k$:  Classified emotion label from $s_{i,k}$ \\
\hspace{0.8em} $R_{i,E}$: Explanation Reward from $o_i$ \\

\textbf{Output}: \\
\hspace{0.8em} $R_E$: List of all computed Explanation Rewards\\
\begin{algorithmic}[1]
\STATE Initialize $R_E \leftarrow [\ ]$
\FOR{$i = 1$ to $G$}
    \STATE Split $E_i$ into sentences $S_i = \{s_{i,1}, s_{i,2}, \dots, s_{i,l}\}$
    \STATE $c_i \leftarrow 0$, $N_{i,neu} \leftarrow 0$
    
    \FOR{each $s_{i,k}$ in $S_i$}
        \STATE $u_{i,k} \leftarrow torch.sigmoid(ERV(s_{i,k}))$
        \IF{$\sum(u_{i,k} > \tau) == 0$}
            \STATE $EL_k \leftarrow$ Top-1 label from $u_{i,k}$
        \ELSIF{$\sum(u_{i,k} > \tau) == 1$}
            \STATE $EL_k \leftarrow$ Label corresponding to the argmax of $u_{i,k}$
        \ELSE
            \STATE $EL_k \leftarrow$ Top-$k$ labels from $u_{i,k}$
        \ENDIF
        \IF{\texttt{"neutral"} $\in EL_k$}
            \STATE $N_{i,neu} \leftarrow N_{i,neu} + 1$
        \ENDIF
        \IF{$e_{gt} \in EL_k$}
            \STATE $c_i \leftarrow c_i + 1$
        \ENDIF
    \ENDFOR

    \IF{$e_{gt} ==$ \texttt{"neutral"}}
        \STATE $R_{i,E} \leftarrow \frac{c_i}{|S|}$
    \ELSE
        \STATE $R_{i,E}
        \leftarrow \frac{c_i}{|S|-N_{i,neu}} $
    \ENDIF\\
    $R_E$.append($R_{i,E}$)
\ENDFOR
\STATE \textbf{return} $R_E$
\end{algorithmic}
\end{algorithm}

\par\vfill\null
\columnbreak
\vspace*{0pt} 

\section{Qualitative Results}
\begin{figure*}[t!]
    \centering
    \includegraphics[width=0.92\textwidth]{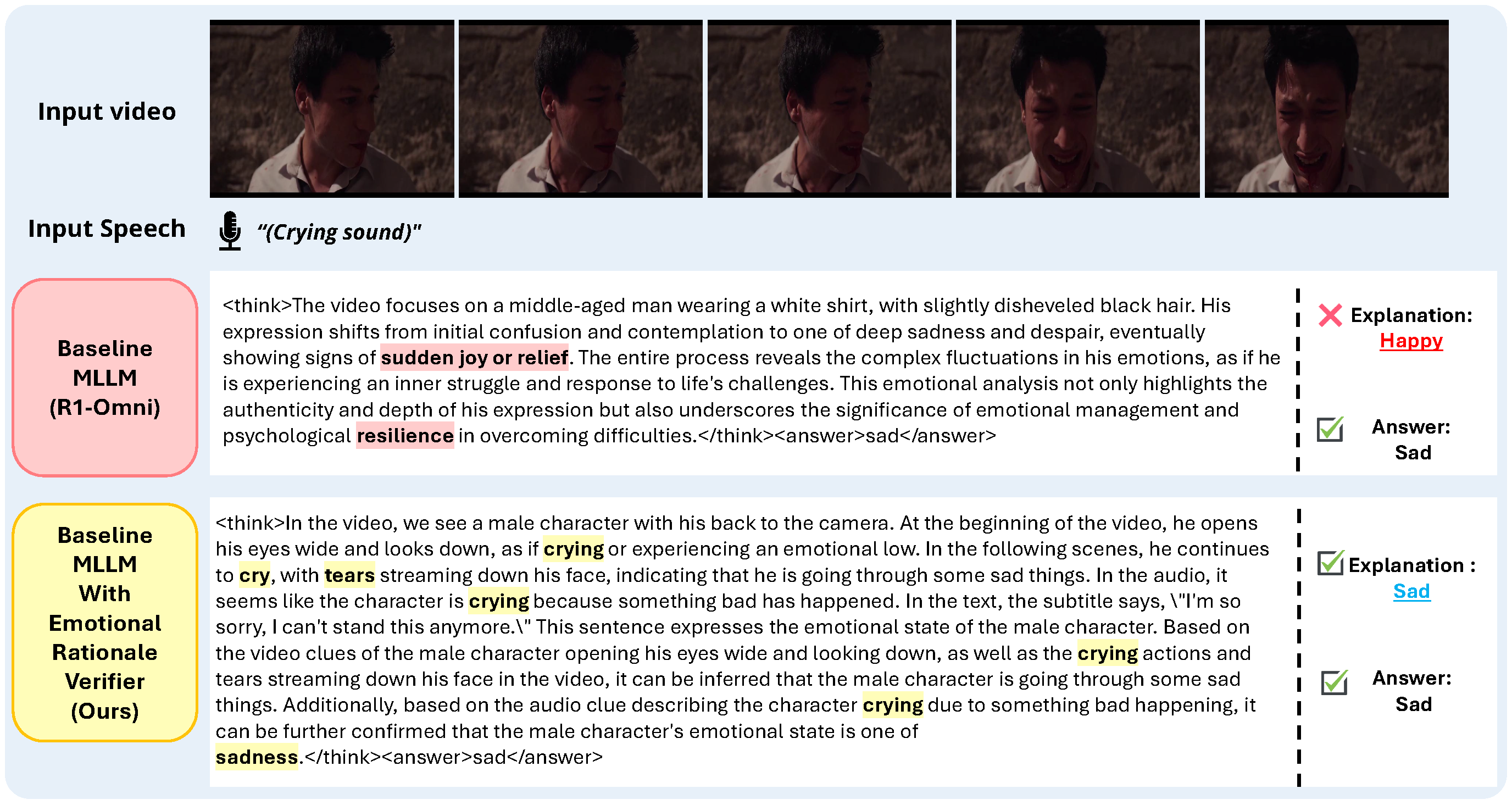}
    \caption{Generated emotional explanation comparison between R1-Omni vs. Ours about 15341.mp4 video in MAFW dataset}
    \label{fig:Qualitative2}
\end{figure*}

\begin{figure*}[t]
    \centering
    \includegraphics[width=0.92\textwidth]{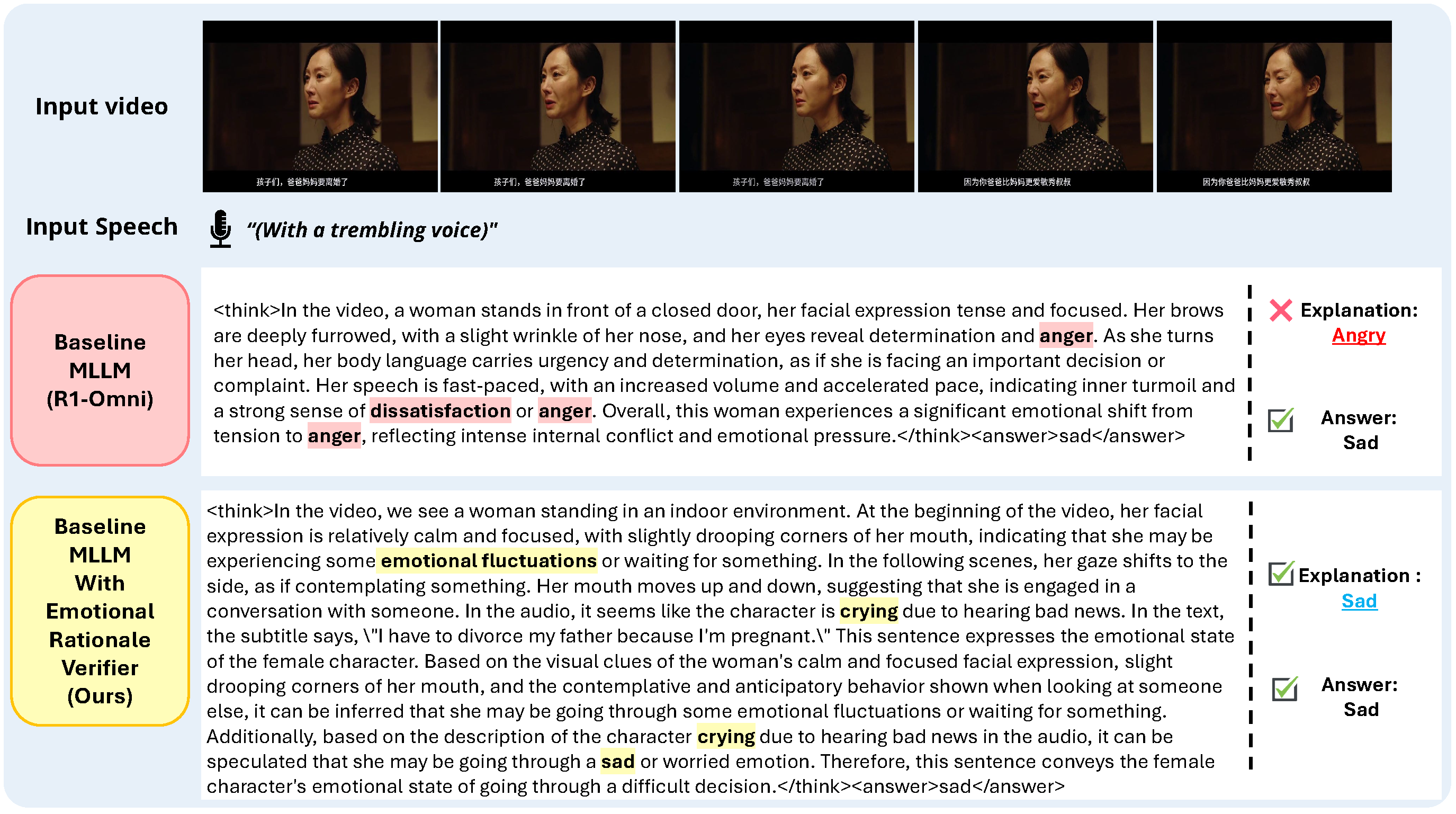}
    \caption{Generated emotional explanation comparison between R1-Omni vs. Ours about 11003.mp4 video in MAFW dataset}
    \label{fig:Qualitative3}
\end{figure*}

\begin{figure*}[t]
    \centering
    \includegraphics[width=0.92\textwidth]{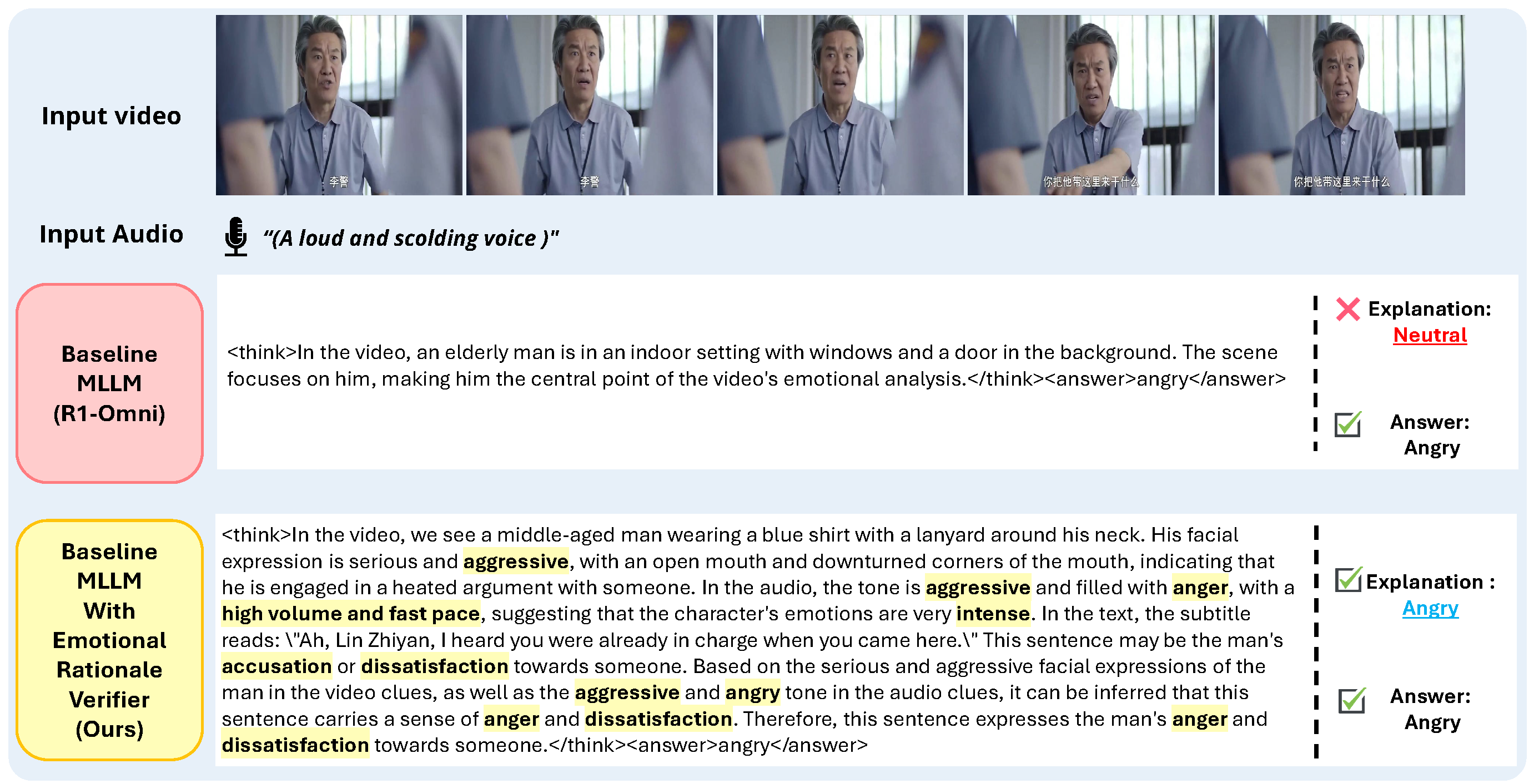}
    \caption{Generated emotional explanation comparison between R1-Omni vs. Ours about 10360.mp4 video in MAFW dataset}
    \label{fig:Qualitative4}
\end{figure*}

\begin{figure*}[t]
    \centering
    \includegraphics[width=0.92\textwidth]{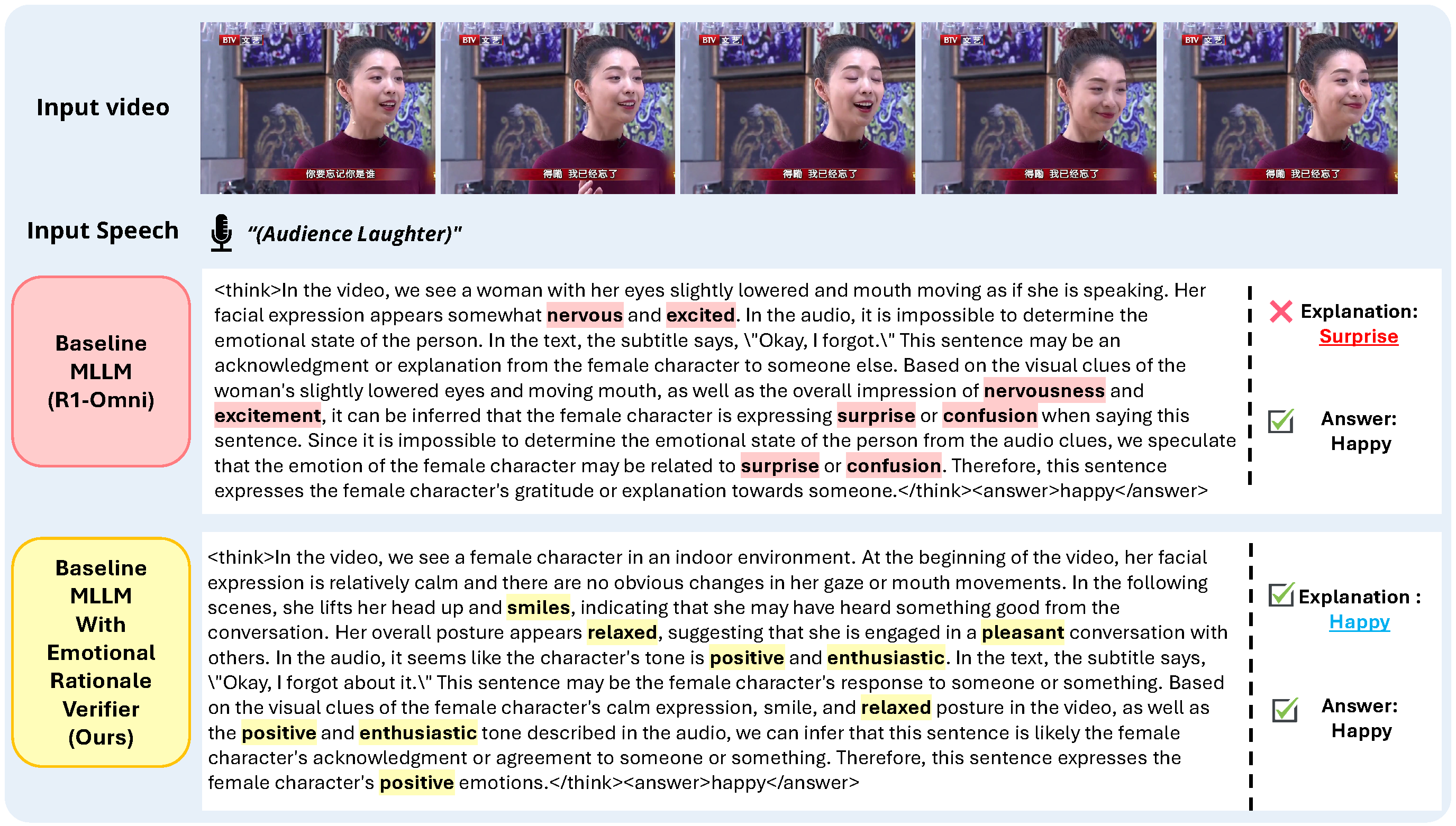}
    \caption{Generated emotional explanation comparison between R1-Omni vs. Ours about 15472.mp4 video in MAFW dataset}
    \label{fig:Qualitative5}
\end{figure*}

Figure~\ref{fig:Qualitative2},~\ref{fig:Qualitative3},~\ref{fig:Qualitative4},~\ref{fig:Qualitative5} show qualitative results comparing the generated explanations from R1-Omni and Ours. While R1-Omni produces the correct ground-truth emotion in the $<$answer$>$ format, its predicted emotions from the closed-source LLM are often misaligned. In contrast, our model’s explanations and final predicted answers demonstrate strong coherence.

\clearpage

\end{document}